# ARDOP: A Versatile Humanoid Robotic Research Platform

Sudarshan S Harithas
BMS College of Engineering
Bangalore, Karnataka, India
1bm16ec109@bmsce.ac.in

Harish V Mekali
BMS College of Engineering
Bangalore, Karnataka, India
hvm.ece@bmsce.ac.in

## ABSTRACT

This paper describes the development of a humanoid robot called ARDOP. The goal of the project is to provide a modular, open-source, and inexpensive humanoid robot that would enable researchers to answer various problems related to robotic manipulation and perception. ARDOP primarily comprises of two functional units namely the perception and manipulation system, here we discuss the conceptualization and design methodology of these systems and proceed to present their performance results on simulation and various custom-designed experiments. The project has been maintained on our page https://ardop.github.io/.

## 1 INTRODUCTION

The multidisciplinary research community has been increasingly focusing on developing robots that can assist and perform activities with humans in the loop. To serve the increase in attention, a robust modular research platform that can aid in rapid testing and software development, while being affordable, simple to reconfigure is of prime importance. Here we present *Autonomous Robot Development Open Source Platform (ARDOP)* a robot that intends to accelerate research primarily in the domains of manipulation [1], and perception [2] [3]. The robot is depicted in Fig. 1

At the core ARDOP consists of two functional units namely the *perception* and *manipulation system* . A calibrated RGBD Microsoft Kinect camera [4] and the pan-tilt joint of the neck form the perception system, the manipulation system includes a pair of 6DOF robotic manipulators and its corresponding control system. A detailed description of the conceptualization, design, and development of individual systems has been presented in this paper. Furthermore, we present the *ARDOP arm* where the manipulator is constructed considering the workspace and torque constraints. Additionally, we evaluate the independent and coordinated performance of these systems through a series of experiments such as *workspace aware object manipulation* and *Tic-Tac-Toe*.

Research across industry and academia have resulted in a few of the finest robots such as PR2 [5] , Robot Cosero [6],Intel HERB[7], Armar III [8], Pepper [9], bipedal robots such as ASIMO [10] , ASIMO [10], and Petman [11], Atlas [12] form another class of humanoid robots. These robots have helped to expedite research but despite their impressive performance and encouraging results they are very expensive for purchase and only a few of them are open source but are difficult to replicate. Whereas all the hardware designs and software packages of ARDOP have been made available open-source and it can be replicated within $1000.

Our contribution can be summarized as follows:

(1) Development of a robust, modular, inexpensive humanoid robot and open-source of its design files and software packages that can be easily reproduced.
(2) Formulation and design of 6DOF robotic manipulator considering workspace and torque constraints. The arm can perform object manipulation to sub-centimeter range accuracy.

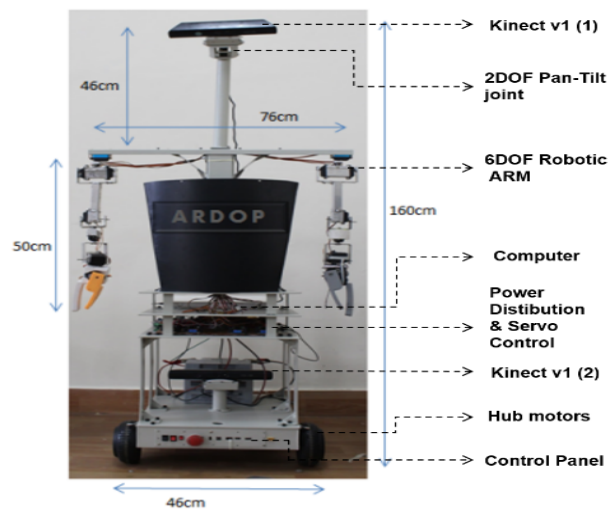

Figure 1: ARDOP, a Humanoid Robotic Research Platform.

## 2 DESIGN

The structural design of ARDOP depicted in Fig. 3, the robot is built using aluminum and is powder coated to ensure electric insulation and enhanced aesthetic beauty. The design of the robot is inspired by the bodily structure of "centaur" [1]. In this section, we describe the mechatronic architecture of ARDOP, followed by the manipulator design and kinematics.

### 2.1 Mechatronic Architecture

The mechatronic architecture describes the interactions between the actuators, sensors, computation, and power distribution circuits. The mechatronic architecture of ARDOP is depicted in Fig. 4.

The architecture includes three modules

- Head: RGBD Camera (Kinect) with pan-tilt neck joint (2DOF actuator).

---
[1] https://en.wikipedia.org/wiki/Centaur

- Arm: a pair of 6DOF manipulators.
- Upper Base: Housing for computer, drivers, power supply units, and controllers.

A 12V 60Ah battery connects to the power distribution and servo control board and powers the manipulation and the perception systems of ARDOP. The power supply is secured and monitored by the safety circuits and power meter respectively. The manipulation system consists of two 6DOF arms which are controlled using the Arduino Mega development board and the perception system consists of a Kinect camera and a 2DOF pan-tilt joint. Furthermore, ARDOP has a tall neck of length $46cm$, which places the camera at an optimal viewing position, additionally, it helps to counter the blind spot in the depth image of the Kinect camera. The laptop with Intel i5-6200U CPU with 16 GB RAM and a 1 TB HDD is housed in the *upper base* performs the computation and inter-system coordination.

## 2.2 Software Architecture

The modular software stack allows for simple user interface and development, it comprises of four layers as depicted in Fig. 2

- *Hardware Abstraction* The layer comprises of the drivers and controllers that communicate with the hardware actuators/sensors.
- The *OS* layer consists of the *Robot Operating System* that is central to the software architecture. This layer provides various tools and libraries that are required for the user level program development.
- *System Service* this layer includes the various user defined programs that can be deployed on the robot. Fig. 2 depicts the program stack of the three systems specific to the various experiments that we had conducted as detailed in section 3.
- *Application* this is an actionable layer where the final implementation of the program can be observed.

## 2.3 Manipulator Design

ARDOP is geometrically symmetric in the sagittal plane. It has two identical 6 DoF arms with its shoulder joint equidistant from the base of the neck. Fig. 5, depicts the geometric and coordinate frame representation of ARDOP. A 3DOF spherical wrist has been used to enable the manipulator to access the complete volume of the workspace in all possible orientations, the ARDOP arm is shown in Fig. 6.

The arm design involves determining the mass, link length and selecting appropriate motors for a given payload. From the stick diagram of the arm is shown in Fig. 6(a), the following considerations are made:

(1) The arm length $L_1 + L_2 + L_{eef} = 50cm$.
(2) The end-effector length $L_{eef} = 8cm$, this implies $L_1 + L_2 = 42cm$.
(3) Maximum payload mass $m_p = 0.2kg$ (inclusive of the mass of the end-effector).
(4) All six motors of the same mass $(m)$, $m_{s0} = m_{s1} = m_{s2} = m_{s3} = m_{s4} = m_{s5} = m = 0.064kg$.

It can be observed from Fig. 6, that the mass of the arm is distributed along the axis of motor $S_0$, which decreases its required torque. Furthermore, we can interpret that motors at joints $S_1$ and $S_2$ would have a higher torque requirement when compared to motors at other joints. Therefore, we consider the torque $T_1$ (of motor $S_1$) and $T_2$ (of motor $S_2$) to constrain and obtain the optimal link length $L_1$ and $L_2$, that would enable the arm to access maximum workspace volume.

Considering Fig. 6(a), the torque $T_1$ and $T_2$ can be obtained from Eq. 1 and 2 respectively, where $m_{L1}$ and $m_{L2}$ is the mass of the link.

$$T_1 = m_{L1}(\frac{L_1}{2}) + m(L_1) + m_{L2}(L_1 + \frac{L_2}{2}) + 3m(L_1 + L_2)$$
$$+ m_p(L_1 + L_2 + L_{eef}) \quad (1)$$

$$T_2 = m_{L2}(\frac{L_2}{2}) + 3m(L_2) + m_p(L_2 + L_{eef}) \quad (2)$$

An aluminium sheet with density $(D) = 2.7 \times 10^{-3} \frac{Kg}{cm^3}$, thickness (H) $0.3cm$ and breadth (B) of $12cm$ is used for arm design. Considering the rectangular structure of the sheet, the volume of link L1 and L2, their volume can be expressed as,

$$V = L_{1,2} \times 12 \times 0.3$$

and mass of link L1 and L2 can be expressed as,

$$Mass(M) = Density(D) \times Volume(V)$$

$$m(L_{1,2}) = 2.7 \times 10^{-3} \times 12 \times 0.3 \times L_{1,2}$$

$$m(L_{1,2}) = 9.72 \times 10^{-3} \times L_{1,2} = K \times L_{1,2} \quad (3)$$

Substituting $L_2 = 42 - L_1$ and the above results in Eq. 1 and 2,

$$T_1 = mL_1 + \frac{k(42^2)}{2} + 3m(42) + 0.2(50) \quad (4)$$

$$T_2 = \frac{K(42 - L1)^2}{2} + 3m(42 - L_1) + 0.2(42 - L_1 + 8) \quad (5)$$

The worst-case torque $(T_{1,2wc})$ analysis is performed by increasing the torque requirement by 75% i.e. $T_{1,2} = 1.75 \times T_{1,2}$ and considering efficiency factor $(\eta = 0.7)$ for the available motor with a stall torque of $35Kg.cm$. Two motors are used in the shoulder joint $S_1$ to increase the stability and robustness of the arm, moreover, this implies that the effective torque at joint $S_1$ is $70Kg.cm$. The worst-case analysis factor $(\beta)$ is given by,

$$\beta = \frac{\eta}{1.75} = \frac{0.7}{1.75} = 0.4$$

Hence, $T_{1wc} = 70\beta = 28Kg.cm$, and $T_{2wc} = 35\beta = 14Kg.cm$.

$$T_1 < T_{1wc} \quad and \quad T_2 < T_{2wc} \quad (6)$$

A graph as shown in Fig. 7 is plotted between the various values of link length $L_1$ and its corresponding expected torque values $T_1$ and $T_2$. For all the points that do not obey the *worst case condition* as given by Eq. 6 are eliminated and a value of "−10" would assigned to them indicating their elimination. From the graph it can be further inferred that the available link length $L_1$ is between $16cm$ to $24cm$. We choose an optimal and safe value of $L_1 = 20cm$ and $L_2 = 42 - L_1 = 22cm$.

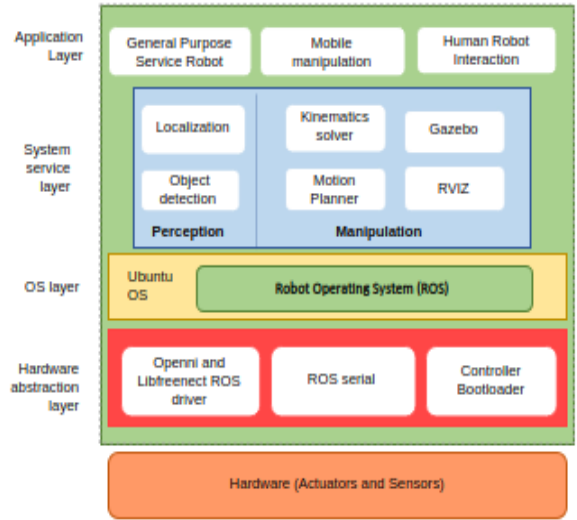

Figure 2: Software Design

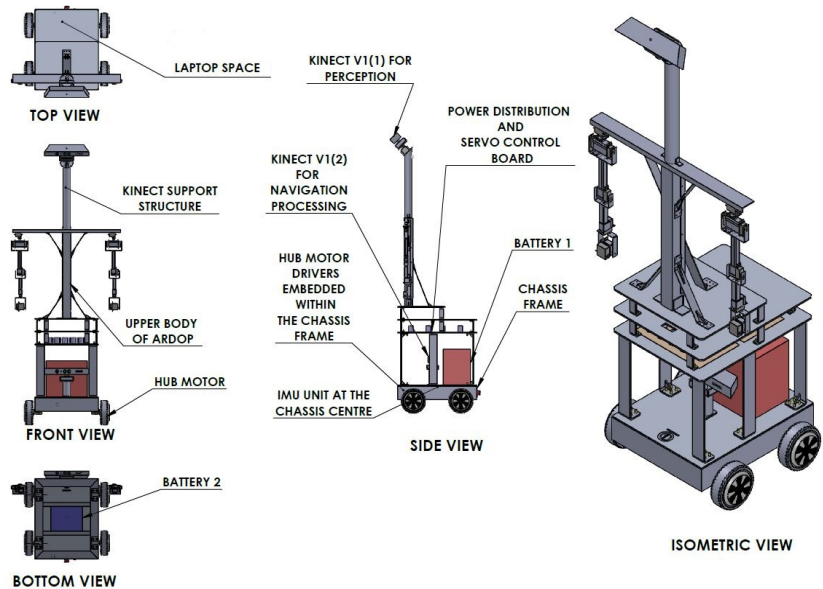

Figure 3: ARDOP Structural Design

2.3.1 **Simulation**. MoveIt! [13] offers *KDL kinematics plugin* [14] along with the *OMPL motion planner* [15] to generate the kinematics solution and the trajectory plan. Rviz [2] [16] is a 3D-visualization tool within ROS [17], it offers interactive GUI to set the end-effector position. The generated kinematic solution is passed to the *Gazebo simulator* [18] through move_group as shown in Fig. 8 (a). Fig. 8 (b) illustrates the implementation of trajectory such that it avoids the obstacle (cube) that is added into the workspace along the expected trajectory towards the goal.

## 2.4 Kinematics

Kinematics of a robotic manipulator involves the development of transformations between the joint-variable space and the cartesian (world coordinate) spaces.

*2.4.1 Forward Kinematics.* Forward kinematics is used to determine the 6DOF pose of an arm given known joint variables. The

---
[2]http://wiki.ros.org/rviz

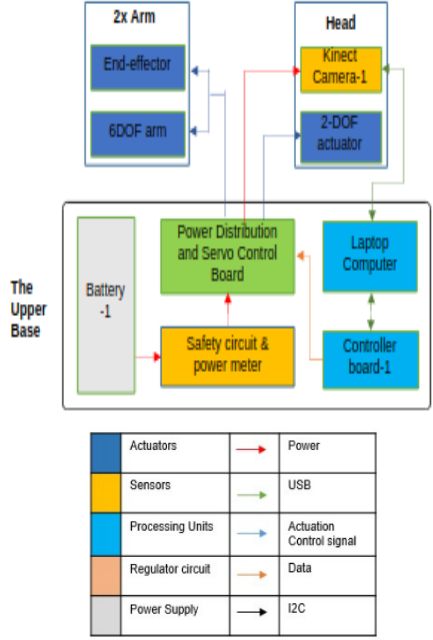

Figure 4: Mechatronic Architecture

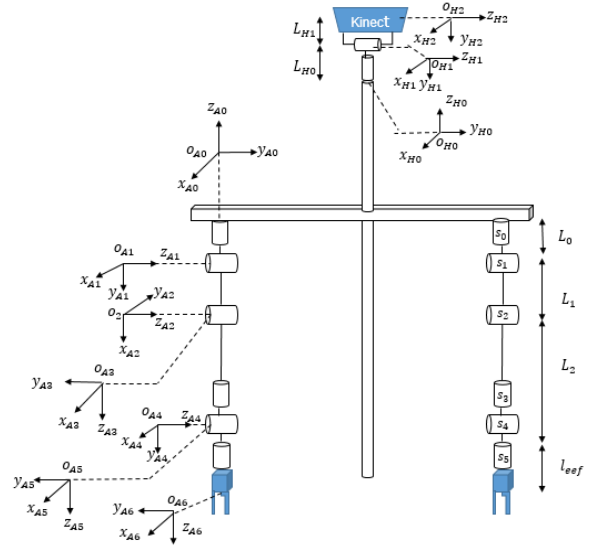

Figure 5: Coordinate frame representation of ARDOP

Denavit-Hartenberg (DH) technique [19] [20] is used to obtain the forward kinematics solution. The DH parameters for the ARDOP manipulator and the head is shown in Table 1(a) and (b) respectively, the corresponding coordinate frame representation has been depicted in Fig. 5

Table 1: DH parameters

(a) Robotic arm

| $\theta$ | $\alpha$ | r | d |
|---|---|---|---|
| $\theta_0$ | $-\pi/2$ | 0 | $-L_0$ |
| $\theta_1 + \pi/2$ | 0 | $L_1$ | 0 |
| $\theta_2 - \pi/2$ | $-\pi/2$ | 0 | 0 |
| $\theta_3$ | $\pi/2$ | 0 | $L_2$ |
| $\theta_4$ | $-\pi/2$ | 0 | 0 |
| $\theta_5$ | 0 | 0 | $l_{eef}$ |

(b) Pan and tilt of head joint.

| $\theta$ | $\alpha$ | r | d |
|---|---|---|---|
| $\theta_{H0}$ | $\pi/2$ | 0 | $L_{H0}(5cm)$ |
| $\theta_{H1}$ | 0 | $L_{H1}(3.4cm)$ | 0 |

$$H_n^{n+1} = \begin{bmatrix} \cos\theta_n & -\sin\theta_n \cos\alpha_n & \sin\theta_n \sin\alpha_n & r_n \cos\theta_n \\ \sin\theta_n & \cos\theta_n \cos\alpha_n & -\cos\theta_n \sin\alpha_n & r_n \sin\theta_n \\ 0 & \sin\alpha_n & \cos\alpha_n & d_n \\ 0 & 0 & 0 & 1 \end{bmatrix}$$

$$H_{H0}^{H2} = H_{H0}^{H1} \cdot H_{H1}^{H2} \quad (7)$$

$$H_{A0}^{A6} = H_{A0}^{A1} \cdot H_{A1}^{A2} \cdot H_{A2}^{A3} \cdot H_{A3}^{A4} \cdot H_{A4}^{A5} \cdot H_{A5}^{A6} \quad (8)$$

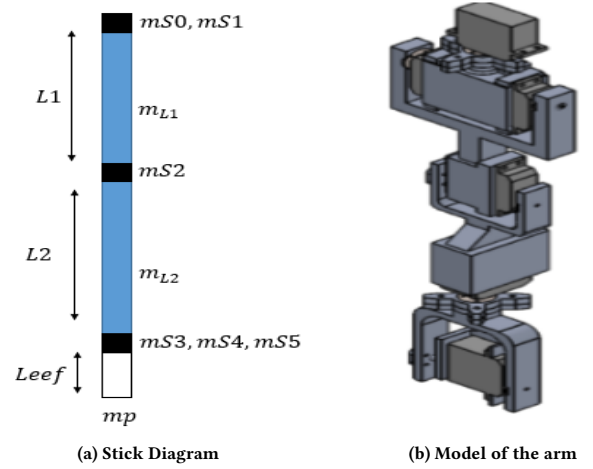

(a) Stick Diagram  (b) Model of the arm

Figure 6: The arm design

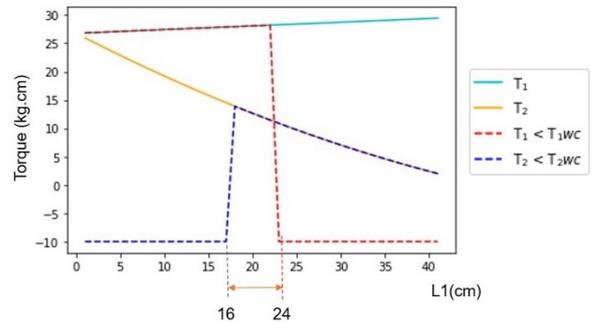

Figure 7: Torque Vs Link Length

Eq. 7 is the homogeneous transform from the camera frame $O_{H2}$ to the neck base frame $O_{H0}$ (this is the extrnsic matrix for section 3.1). The transform from the arm base frame $O_{A0}$ at shoulder joint($S_0$) to the end-effector frame $O_{A6}$ is obtained using Eq. 8. Transform from neck base to shoulder base can be inferred from Fig. 1 and Fig. 5 as $[Z_{H0} - 46]$ with $[Y_{H0} - (76/2)]$ for right arm and $[Y_{H0} + (76/2)]$.

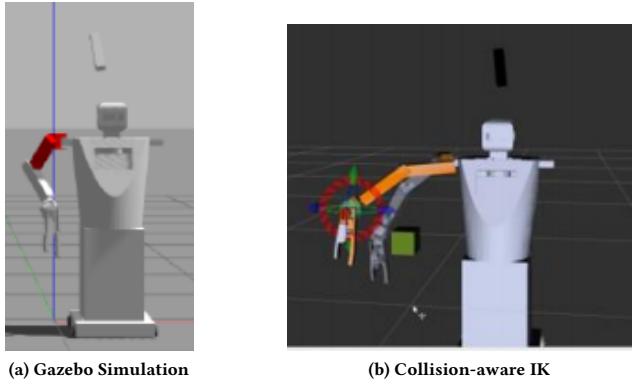

(a) Gazebo Simulation  (b) Collision-aware IK

Figure 8: Gazebo and Rviz simulation of the manipulation system.

*2.4.2 Inverse Kinematics(IK).* The problem of determining the joint angles given the configuration of the end effector is known as inverse kinematics [20]. The pseudo-inverse Jacobian method is used to solve the inverse kinematics of the manipulator. IK is solved in two steps, firstly, we determine the joint angles that effect the position of the end-effector and secondly we solve for the joint angles of the spherical wrist that effect the orientation of the end-effector.

(1) **Solution for the Position of the End-Effector:**
The IK solution for the first three joint angles $\theta_0, \theta_1, \theta_2$, determine the position of the end-effector. From Fig. 9(a) the $'y'$ and $'z'$ are obtained as given in Eq. 10 and 11 and from Fig. 9(b) $'x'$ is determined as given in Eq. 9. The equations 9, 10, and 11 describe the position of the arm in space as a function of input joint angles.

$$x = (L_1 \sin \theta_1 + (L_2 + l_{eef}) \sin(\theta_1 + \theta_2)) \sin \theta_0 \quad (9)$$

$$y = (L_1 \sin \theta_1 + (L_2 + l_{eef}) \sin(\theta_1 + \theta_2)) \cos \theta_0 \quad (10)$$

$$z = L_1 \cos \theta_1 + (L_2 + l_{eef}) \cos(\theta_1 + \theta_2) + L_0 \quad (11)$$

$$X = f(\theta) \quad (12)$$

$$X = [x, y, z] \quad and \quad \theta = [\theta_0, \theta_1, \theta_2]$$

The Jacobian method attempts to minimize the error between the current position ($X_n$) and the final position ($X_d$) of the end effector. The error is given by

$$\Delta X_n = X_d - X_n \quad (13)$$

An incremental change $\Delta \theta_n$ is made in a direction such that the equations converge and the solution is obtained.

$$\Delta \theta_n = [J(\theta)]^{-1} . \Delta X_n \quad (14)$$

$$\theta_{n+1} = \theta_n + \gamma \Delta \theta_n \quad (15)$$

Where $\gamma$ is the step size parameter which determines the rate of convergence and $J(\theta)$ is the Jacobian matrix.

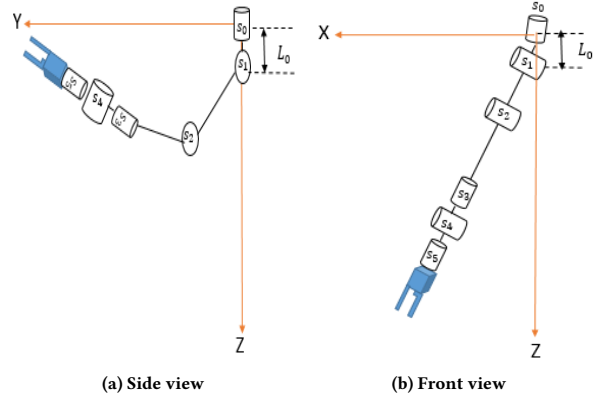

(a) Side view  (b) Front view

Figure 9: Projection of the arm on vertical and profile plane.

(2) **Solution for the Orientation of the End-Effector:** The results from the previous computation of $\theta_0, \theta_1, \theta_2$ is used to obtain the joint angles of the spherical wrist $\theta_3, \theta_4, \theta_5$ that decides orientation of the end-effector. The rotation matrix from the arm base frame to the end-effector is expressed as,

$$R_0^6 = R_0^1 R_1^2 R_2^3 R_3^4 R_4^5 R_5^6 \quad => \quad R_0^6 = R_0^3 R_3^6 \quad (16)$$

From Eq. 16, matrix $R_0^3$ can be obtained by substituting for $\theta_0, \theta_1, \theta_2$. The matrix $R_3^6$ is the rotation matrix of the 3DOF-spherical wrist and orientation of the end-effector.

$$R_3^6 = [R_0^3]^{-1} R_0^6 \quad (17)$$

Eq. 17 is solved to obtain the solution for $\theta_3, \theta_4, \theta_5$.

## 3 EXPERIMENTS AND RESULTS

Experiments were conducted to evaluate the performance of individual systems and to demonstrate a few tasks the robot can perform as a whole. The experiments were *object localization*, *workspace aware manipulation*, *motion planning* and *Tic-Tac-Toe*, this section describes the conduction and results of these tests.

### 3.1 Object Localization

YOLOv3 [21] trained on the COCO dataset [22] is used to detect the objects of interest in space. Object class, bounding box, confidence are the outputs of the YOLOv3 model and the centroid of the bounding box is used as the corresponding point of reference for the object.

This experiment measures the accuracy with which the point of interest can be localized in space. The experimental setup is shown in Fig. 10, the task was to determine the dimension of the rectangular strip given the known pixel co-ordinates of the corners.

The perspective camera projection model [23] [2] was used to recover the 3D location of the corners. The ROS camera calibration [3] is used to determine the camera intrinsic matrix **K**. Using the pixel coordinates $\mathbf{p} = (u, v)^\top$ and the depth map $D$, the following equation 18 is used to determine the 3D location **P** of the object of interest in the neck-base coordinate frame

$$\tilde{\mathbf{P}} = D(\mathbf{p})\mathbf{H}_{oH0}^{oH2^{-1}}\mathbf{K}^{-1}\tilde{\mathbf{p}} \qquad (18)$$

where $D(\mathbf{p})$ is the depth value at the pixel coordinates **p**, $\mathbf{H}_{oH0}^{oH2}$ is the homogeneous transformation between the neck base frame and the camera coordinate frame, and $\tilde{\mathbf{p}}$ represents point in its homogeneous form. Multiple trails were performed changing the position and orientation of the strip and the system was capable of determining the location of the point within a localization error of 0.4$cm$.

---
[3]http://wiki.ros.org/openni_launch/Tutorials/IntrinsicCalibration

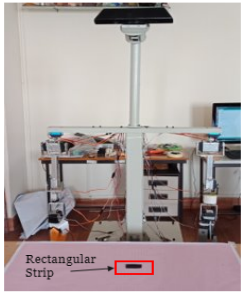

Figure 10: Experimental setup

## 3.2 Workspace Aware Manipulation

This experiment involves to pick and place a ball of radius 3$cm$ and drop it into a cup of radius 3.5$cm$, this experiment validates the arm design and the kinematics formulation. Furthermore, it demonstrates the robustness of the arm to access the points that are at the edge of the workspace.

The experimental setup is shown in Fig. 11 where the ball and cup were placed in the workspace of the left and right arm respectively Fig. 11(a). The manipulation task was completed by performing a handover operation where the left arm picks up the ball Fig. 11(c) and places it within the reach of the right arm Fig. 11(d). Next, the right arm continues the task to pick up Fig. 11(e) and drop the object into the cup Fig. 11(f).

Over 300 trials were conducted, this includes about 100 pick and place experimental trail and over 200 measurement trail to assess the accuracy of the manipulation system. The error in the manipulation system was determined to be : $e_x < 0.1cm$; $e_y < 0.7cm$; $e_z < 1cm$. where $e_x, e_y, e_z$ are the maximum value of obtained error in the 'x' , 'y' , 'z' direction (with reference to Fig. 9) respectively.

## 3.3 Human Robot Interaction

A game of *Tic-Tac-Toe* has been conducted to demonstrate the ability of ARDOP to respond to a given situation in real time and to coordinate with humans. The experimental setup shown in Fig. 12, consists of a rectangular game board of dimension 29.7$cm$ × 42.0$cm$ and each cell was of size 13$cm$ × 9$cm$. Token "O" is a circle of 3.5$cm$ radius used by the human and for ARDOP is assigned token X which was cut from a square of length 3$cm$. In Fig. 12(a) the move made by the human is detected and the logical response is computed, ARDOP also collects the "X" token from the human participant. The computed response is executed by the manipulation system where the collected X token is placed as depicted in Fig. 12 (b). Over 20 iterations, the results were found to be satisfactory without errors.

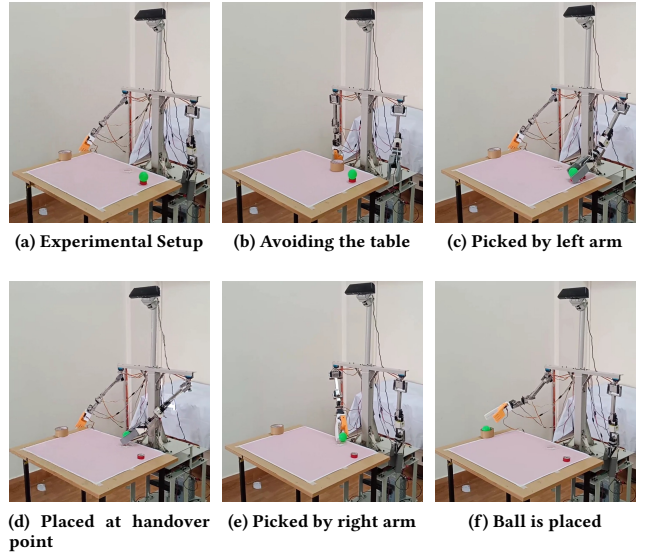

(a) Experimental Setup  (b) Avoiding the table  (c) Picked by left arm

(d) Placed at handover point  (e) Picked by right arm  (f) Ball is placed

Figure 11: Workspace-aware manipulation system.

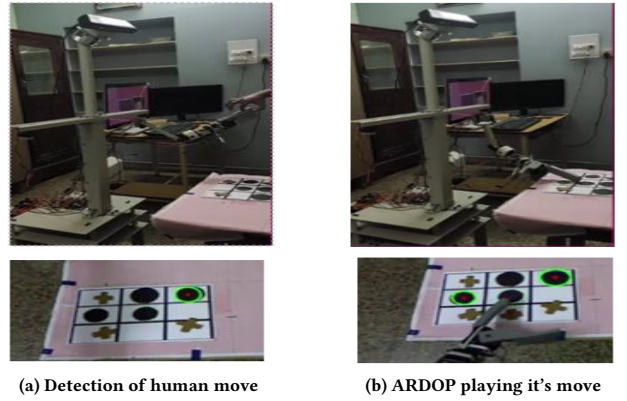

(a) Detection of human move  (b) ARDOP playing it's move

Figure 12: Implementation of Tic-Tac-Toe

## 4 CONCLUSION

ARDOP has been successfully designed and constructed, the development and integration of individual sub-systems has been described in this paper. Furthermore, we describe the design of the ARDOP arm considering the torque and workspace constrains and demonstrate through the *workspace aware manipulation* experiment that the perception system can localize objects and the arm can perform object manipulation with their respective accuracies in the sub-centimeter range. Additionally, the *Tic Tac Toe* experiment demonstrates the human-robot interaction and real time response abilities of ARDOP.

# REFERENCES

[1] Richard P Paul. *Robot manipulators: mathematics, programming, and control: the computer control of robot manipulators.* Richard Paul, 1981.

[2] Yi Ma, Stefano Soatto, Jana Kosecka, and S Shankar Sastry. *An invitation to 3-d vision: from images to geometric models*, volume 26. Springer Science & Business Media, 2012.

[3] Jian Chen, Bingxi Jia, and Kaixiang Zhang. *Multi-View Geometry Based Visual Perception and Control of Robotic Systems.* CRC Press, 2018.

[4] R. A. Newcombe, S. Izadi, O. Hilliges, D. Molyneaux, D. Kim, A. J. Davison, P. Kohi, J. Shotton, S. Hodges, and A. Fitzgibbon. Kinectfusion: Real-time dense surface mapping and tracking. In *2011 10th IEEE International Symposium on Mixed and Augmented Reality*, pages 127–136, 2011.

[5] Quigley, M., Conley, K., Gerkey, B., Faust, J., Foote, T., Leibs, J., et al. (2009). "ros: An open-source robot operating system". *ICRA Workshop on Open Source Software, Kobe.*

[6] Jörg Stückler, Max Schwarz and Sven Behnke, Institute for Computer Science VI, Autonomous Intelligent Systems, University of Bonn,Bonn, Germany. Mobile manipulation, tool use, and intuitive interaction for cognitive service robot cosero. *Frontiers in Robotics and AI*, TECHNOLOGY REPORT ARTICLE, 07 November 2016.

[7] Srinivasa, S.S., Ferguson, D., Helfrich, C.J. et al. Herb: a home exploring robotic butler. *Auton Robot 28, 5 (2010).*

[8] T. Asfour, K. Regenstein, P. Azad, J. Schroder, A. Bierbaum, N. Vahrenkamp, and R. Dillmann. Armar-iii: An integrated humanoid platform for sensory-motor control. In *2006 6th IEEE-RAS International Conference on Humanoid Robots*, pages 169–175, 2006.

[9] Amit Kumar Pandey and Rodolphe Gelin. A mass-produced sociable humanoid robot: Pepper: The first machine of its kind. *IEEE Robotics & Automation Magazine*, 25(3):40–48, 2018.

[10] Yoshiaki Sakagami, Ryujin Watanabe, Chiaki Aoyama, Shinichi Matsunaga, Nobuo Higaki, and Kikuo Fujimura. The intelligent asimo: System overview and integration. In *IEEE/RSJ international conference on intelligent robots and systems*, volume 3, pages 2478–2483. IEEE, 2002.

[11] Gabe Nelson, Aaron Saunders, Neil Neville, Ben Swilling, Joe Bondaryk, Devin Billings, Chris Lee, Robert Playter, and Marc Raibert. Petman: A humanoid robot for testing chemical protective clothing. *Journal of the Robotics Society of Japan*, 30(4):372–377, 2012.

[12] Siyuan Feng, X Xinjilefu, Christopher G Atkeson, and Joohyung Kim. Optimization based controller design and implementation for the atlas robot in the darpa robotics challenge finals. In *2015 IEEE-RAS 15th International Conference on Humanoid Robots (Humanoids)*, pages 1028–1035. IEEE, 2015.

[13] David Coleman, Ioan Alexandru Sucan, Sachin Chitta, and Nikolaus Correll. Reducing the barrier to entry of complex robotic software: a moveit! case study. *CoRR*, abs/1404.3785, 2014.

[14] Herman Bruyninckx, Peter Soetens, and Bob Koninckx. The real-time motion control core of the Orocos project. In *IEEE International Conference on Robotics and Automation*, pages 2766–2771, 2003.

[15] Mark Moll, Ioan A. Şucan, and Lydia E. Kavraki. Benchmarking motion planning algorithms: An extensible infrastructure for analysis and visualization. *IEEE Robotics & Automation Magazine*, 22(3):96–102, September 2015.

[16] Hyeong Ryeol Kam, Sung-Ho Lee, Taejung Park, and Chang-Hun Kim. Rviz: a toolkit for real domain data visualization. *Telecommunication Systems*, 60(2):337–345, 2015.

[17] Anis Koubâa. *Robot Operating System (ROS).*, volume 1. Springer, 2019.

[18] Nathan Koenig and Andrew Howard. Design and use paradigms for gazebo, an open-source multi-robot simulator. In *IEEE/RSJ International Conference on Intelligent Robots and Systems*, pages 2149–2154, Sendai, Japan, Sep 2004.

[19] Chapter 3 forward kinematics: The denavit-hartenberg convention. Duke university.

[20] Mark W Spong, Seth Hutchinson, Mathukumalli Vidyasagar, et al. *Robot modeling and control.* 2006.

[21] Joseph Redmon and Ali Farhadi. Yolov3: An incremental improvement. *CoRR*, abs/1804.02767, 2018.

[22] Tsung-Yi Lin, Michael Maire, Serge J. Belongie, Lubomir D. Bourdev, Ross B. Girshick, James Hays, Pietro Perona, Deva Ramanan, Piotr Dollár, and C. Lawrence Zitnick. Microsoft COCO: common objects in context. *CoRR*, abs/1405.0312, 2014.

[23] Robert Collins. Lecture 13: Camera projection ii reading: Tv section 2.4. CSE486, Penn State.